\documentclass[12pt,a4paper]{article}

\usepackage[cmex10,intlimits]{amsmath}
\usepackage{color}
\usepackage{a4wide}
\usepackage{rotating}
\usepackage{cite}
\usepackage{tikz}
\usepackage{bm}
\usepackage{tipa}
\usepackage[utf8]{inputenc}
\usepackage[T1]{fontenc}

\usepackage{hyperref}       % hyperlinks
\usepackage{url}            % simple URL typesetting
\usepackage{booktabs}       % professional-quality tables
\usepackage{amsfonts}       % blackboard math symbols
\usepackage{nicefrac}       % compact symbols for 1/2, etc.
\usepackage{microtype}      % microtypography
\usepackage[intlimits]{amsmath}
\usepackage{amssymb}
\usepackage{mathptmx}

\usepackage{subcaption}
\usepackage{graphicx}

\makeatletter
\def\blfootnote{\xdef\@thefnmark{}\@footnotetext}
\makeatother

\title{A Survey of Learning Curves with Bad Behavior:\\ or\\ How More Data Need Not Lead to Better Performance}
\author{Marco~Loog$^{1,2}$ \hfill Tom J Viering$^1$ \medskip \small \\
\small
\begin{tabular}{c}
$^1$Delft University of Technology, The Netherlands\\
$^2s$University of Copenhagen, Denmark 
\end{tabular}}

\begin{document}

\maketitle

\begin{abstract}
Plotting a learner's generalization performance against the training set size results in a so-called learning curve.  This tool, providing insight in the behavior of the learner, is also practically valuable for model selection, predicting the effect of more training data, and reducing the computational complexity of training.  We set out to make the (ideal) learning curve concept precise and briefly discuss the aforementioned usages of such curves.  The larger part of this survey's focus, however, is on learning curves that show that more data does not necessarily leads to better generalization performance.  A result that seems surprising to many researchers in the field of artificial intelligence. We point out the significance of these findings and conclude our survey with an overview and discussion of open problems in this area that warrant further theoretical and empirical investigation. \medskip 
\end{abstract}

\newpage

%\tableofcontents

\section{Introduction}

A curve that shows the dependence of a learner’s generalization performance as a function of the training set size is known as a learning curve.\footnote{Arguably, the term sample complexity curve could be used as well (cf.~\cite{zubek2016complexity}).}  No matter the problem under consideration, it seems reasonable to require from a learner that its corresponding learning curve behaves monotonically, i.e., the learner delivers improved generalization performance with the availability of increasing amounts of training data.  As any single draw of additional data from the problem at hand could be arbitrary bad by chance, this improved performance should be considered in expectation, or at least averaged over a large number of independent instantiations.  In that respect, the requirement of performance improvement with more data may appear a mere theoretical one: in a real-world scenario, only a finite sample is available to train, tune, and validate a learner and we may not be able to effectively estimate its expected performance.  Nevertheless, also from a practical point of view it is arguably of use to know that, upon collecting more data, one's learner is at least not expected to become worse.

In our experience, partly anecdotal of course, most researchers in AI indeed expect improved performance of their learner with more data.  Actual evidence of this can be found in the literature. \cite{shalev2014understanding}, for instance, states that the learning curve must start decreasing once the training set size becomes larger than the VC-dimension, while \cite{Duda2012} claims that for many real-world problems its decay is monotonically.  \cite{Tax2008} and \cite{Gu2001} state that it is expected that performance improves with more data, \cite{Ting2017} takes it as conventional wisdom that the learning curve acts monotonically, and \cite{Boonyanunta2004} considers this behavior to be widely accepted. \cite{Provost1999} assumes well-behaved curves, which, as a rule, means they are smooth and monotonic \cite{weiss2014generating}.  Moreover, \cite{Amari1992} explicitly states that the generalization error decreases as training set size increases.  In the meta-review of our 2019 paper \cite{loog2019minimizers}---covered later on, this monotonic behavior is tellingly qualified as folk wisdom \cite{anonym}.  Indeed, it may come as a surprise to many researchers that a learner's performance can actually deteriorate with more training data. Even in expectation.

This review goes through different settings in which such unexpected phenomenon can occur and highlights various of its facets.  We refer to these learners as badly behaved or ill-behaved and, by association, we refer to their learning curves in a likewise manner.  Next to providing an in-depth review of the literature related to this particular subject, it also discusses our current understanding of this behavior and provides intriguing open questions and interesting directions for further research.  To keep the survey self-contained, we spend some paragraphs on properly defining (idealized) learning curves.  Subsequently, to also underline the practical relevance of the study of learning curves, we briefly cover its main usages.

A comprehensive review on learning curves can be found in \cite{viering2021shape}.  The current work is inspired by Section 6 from this review and offers a fully revised extension and update of this important section.  For a review complementary to \cite{viering2021shape}, one that delves specifically into how such curves can provide insight into the learning phase and how they are important for meta-learning, we refer to \cite{mohr2022learning}.  Finally, please also check \url{https://github.com/tomviering/ill-behaved-learning-curves}.

\subsection{Outline}

In this work, we review learning curves in the context of standard supervised learning problems such as classification and regression.  Section \ref{sect:def} makes precise what we mean by a learner, a learning problem, and the associated learning curve.  To underline the practical use of learning curves, Section \ref{sect:use} sketches the  insight  into  model  selection  they  can give us, and how they are employed, for instance, in meta-learning  and  reducing  the  cost  of  labeling  or computation.  Sections \ref{sect:bad} and \ref{sect:pabad} then follow with an overview of important cases of learning curves that do not behave well.  The former section focuses on standard learning curves, while the latter considers a particular kind of learning curves that is often used in Bayesian analysis. These two section shows that our understanding of the behavior of learners is more limited than one might expect.  Section \ref{sect:fix} turns to some specific and general approaches to make learners (more) well-behaved, while Section \ref{sect:disc} provides a discussion and concludes our review.

\section{Problems, Learners, and Expected Curves} \label{sect:def}

To formalize learning problems, learners, and learning curves, let u start by introducing $\mathcal{X}$ to denote a generic input space and $\mathcal{Y}$ an output space.  Let $S_N$ indicate a training set of size $N$, which takes input-output pairs from $\mathcal{X} \times \mathcal{Y}$ and acts as input to our learning algorithm $A$, or learner for short. The $N$ pairs $(x_i,y_i)$ of the training set are i.i.d. samples from an unknown probability distribution $P_{XY}$ over $\mathcal{X} \times \mathcal{Y}$, which defines our learning problem.  A trained learner $A(S_N)$ delivers a hypothesis $h$ from some hypothesis class $\mathcal{H}$, which contains all models that can, a priori, be returned by the learner $A$. 

An example of a hypothesis class is the set of all linear models $\{h:x\mapsto a^Tx + b\mid a\in\mathbb{R}^d,b\in\mathbb{R}\}$.  In standard classification and regression, $S_N$ consists of $(x,y)$ pairs, where $x \in \mathbb{R}^d$ is the $d$-dimensional input vector (i.e., the features, measurements, or covariates) and $y$ is the corresponding output (e.g. a discrete class label or continuous regression target).  

When evaluating $h$ on an input $x$, its prediction for the corresponding $y$ is given by $\hat{y} = h(x) \in \mathcal{Y}$. The performance of a particular hypothesis $h$ is measured by a loss function $\ell$ that compares $y$ to $\hat{y}$. Examples are the squared loss for regression, where $\mathcal{Y} \subset \mathbb{R}$ and $\ell_\text{squared}(y,\hat{y}) = \ell(y,\hat{y}) = (y-\hat{y})^2$, and the zero-one loss for (binary) classification, where $\ell_\text{0-1}(y,\hat{y}) = \ell(y,\hat{y}) = \tfrac{1}{2}(1-y\hat{y})$ when $\mathcal{Y}=\{-1,+1\}$.

The standard objective is that our hypothesis performs well on average on all new and unseen observations. Ideally, this is measured by the expected loss or risk $R$ over the true distribution $P_{XY}$:
\begin{equation}\label{eq_risk}
    R(h) = \int \ell(y,h(x)) dP(x,y). 
\end{equation}
Here, as in most that follows, we omit the subscript $XY$ to $P_{XY}$.  Let us already note, in addition, that the evaluation loss employed in Equation \eqref{eq_risk} does not have to match the loss that is considered by the learner $A$ at training time. In fact, in classification, the loss actually optimized is typically different from the 0-1 loss or accuracy that is considered at evaluation and test time (cf. \cite{loog2016measuring}), an issue we return to in Section \ref{sect:dip}.

Now, an \emph{individual} learning curve considers a single training set $S_N$ for every $N$ and calculates its corresponding risk $R(A(S_N))$ as a function of $N$. However, as noted already, a single $S_N$ may deviate significantly from the expected behavior. Therefore, we are often interested in an averaging over many different random draws $S_N$. Ideally, we would like to evaluate the expectation 
\begin{equation}
	\bar{R}_N(A) = \underset{S_N \sim P^N}{\mathbb{E}} R(A(S_N)). \label{eq_exp_lc}
\end{equation}
The plot of $\bar{R}_N(A)$ against the training set size $N$ gives us the expected learning curve. From this point onward, when we mention the term ``learning curve'' without any further specifications, this is what is referred to.

The preceding learning curve is defined for a single learning problem $P$. Sometimes we wish to study how a model performs over a set or range of problems or, more generally, a full distribution $\mathcal{P}$ over such problems.  The learning curve that considers such averaged performance is referred to as the problem-average (PA) learning curve:
\begin{equation}\label{eq:pa}
\bar{R}^\text{PA}_N(A) = \underset{P \sim \mathcal{P}}{\mathbb{E}}  \bar{R}_N(A).
\end{equation}
The general term problem-average was coined in \cite{Duda2012}. PA learning curves are particularly useful in the analysis of Bayesian approaches, where an assumed prior over possible problems often arises naturally. This particular notation of learning curve is primarily employed in Section \ref{sect:pabad}. In the Bayesian literature, the risk integrated over the prior is also called the Bayes risk, integrated risk, or preposterior risk \cite[page~195]{Murphy2012}.  The term preposterior signifies that, in principle, we can determine this quantity without observing any data, as the prior $\mathcal{P}$ prespecifies what data we expect to see.

\section{General Practical Usage}\label{sect:use}

Studies of learning curves have both practical and theoretical value. Here, we do not necessarily make a very strict separation between the two, though the primary emphasis in this survey is on the latter. This section, however, focuses in part of the former and briefly covers the current, most important uses of learning curves when it comes to applications.  A more extensive and complete overview can be found in \cite{mohr2022learning}.  We return to the practical value of the theoretical findings presented in the discussion.

\subsection{Model Evaluation}

Machine learning as a field has shifted more and more to benchmarking learning algorithms. In the last 20 years, for instance, more than 5000 benchmark datasets have been created (see \url{https://paperswithcode.com/} for an overview). These benchmarks are often set up as competitions \cite{sculley2018winner} and investigate which algorithms are better or which novel procedure outperforms existing ones \cite{Perlich2003}.  Typically, a single number, summarizing performance, is used as evaluation measure.

A recent meta-analysis indicates that the most popular measures are accuracy, the F-measure, and precision \cite{blagec2020critical}. 
An essential issue these metrics ignore is that sample size can have a large influence on the relative ranking of different learning algorithms. In a plot of the learning curves of the different learners this would be visible as a crossing of their curves. In that light, it is beneficial if benchmarks consider multiple sample sizes, to get a better picture of the strengths and weaknesses of the approaches. The learning curve provides a concise picture of this sample-size dependent behavior.

Also using learning curves, \cite{Perlich2003} finds that, besides sample size, separability of the problem can be an indicator of which algorithm will dominate the other in terms of the learning curve. Beyond that, the learning curve, when plotted together with the training error of the algorithm can be used to detect whether a learner is overfitting \cite{jain2000statistical,duin2005stat,duin2016pattern,loog2018supervised}.
% Besides sample size, dimensionality seems also an important factor to determine whether linear or non-linear methods will dominate \cite{strang2018don}. To that end, learning curves combined with feature curves may offer further insights. %

\subsection{Reduction of Data Collection Costs}\label{sec:extrapolate}

When collecting data is difficult, time-consuming, or otherwise expensive, the possibility to accurately extrapolate a learner's learning curve can be useful. Extrapolations, which are typically base on some parametric learning curve model, give an impression beforehand of how many examples to collect to come to a specific performance \cite{Frey1999}. As such, they also allows one to judge when data collection can be stopped. Examples of such practice can, for instance, be found in machine translation \cite{Kolachina2012} and medical applications \cite{Mukherjee2003,Hess2010,Figueroa2012}.   \cite{Last2007} quantifies potential savings assuming a fixed cost per collected sample and per generalization error. Extrapolating the learning curve using some labeled data, the point at which it is not worth anymore to label more can be determined and data collection can be stopped.

\subsection{Speeding Up Training, Tuning, and Selecting}\label{sec:extrapolate_comp}

Learning curves can be used to reduce computation time and memory with regards to training models, model selection and hyperparameter tuning. %

Reference \cite{Provost1999} speeds up training by so-called progressive sampling, using a learning curve to determine if less training data can reach adequate performance. If the slope of the curve becomes too flat, learning is stopped, making training potentially much faster. They recommended to use a geometric series for $N$ to reduce computational complexity. 

Several variations on progressive sampling exist. \cite{John1996} proposes the notion of probably close enough where a power-law fit is used to determine if the learner is so-called epsilon-close to the asymptotic performance. \cite{Meek2002} gives a rigorous decision theoretic treatment of the topic. By assigning costs to computation times and performances, they estimate what should be done to minimize the expected costs. Progressive sampling also has been adapted to the setting of active learning \cite{Tomanek2008}. \cite{Leite2004} combines meta-learning with progressive sampling to obtain a further speedup in the training phase.

Model selection can also be sped up. \cite{Leite2005} compares the learners' initial learning curves to a database of learning curves to predict which of two classifiers will perform best on a new dataset. This can be used to avoid costly evaluations using cross validation. Moreover, \cite{Leite2007} proposes an iterative process that predicts the required sample sizes, builds learning curves, and updates the performance estimates in order to compare two classifiers. \cite{VanRijn2015} extends the technique to rank many machine learning models according to their predicted performance, tuning their approach to come to an acceptable answer in as little time as possible.   

With regards to hyperparameter tuning, already in 1994, \cite{Cortes1994} devised an extrapolation scheme for learning curves, based on the fitting of power laws, to determine if it is worth to fully train a neural network. In the deep learning era, this has received renewed attention. \cite{Hestness2017} uses extrapolation to optimize hyperparameters. \cite{hoiem2021learning} takes this a step further and actually optimize several design choices, such as data augmentation. %
For such applications, a good learning curve model is essential.

\section{Badly Behaving Learning Curves}\label{sect:bad}

It is important to understand that learning curves do not always behave well and that this is not necessarily an artifact of the finite sample or the way an experiment is designed.  Deterioration with more training data can obviously occur when considering the individual curve $R(A(S_N))$ for a particular training set, because for every $N$, we can be unlucky with our draw $S_N$. That ill-behavior can also occur in expectation, i.e., for $\bar{R}_N(A)$, however, is less obvious.

\begin{figure}[tbh]
 	\centering
 	\includegraphics[width=0.9\textwidth]{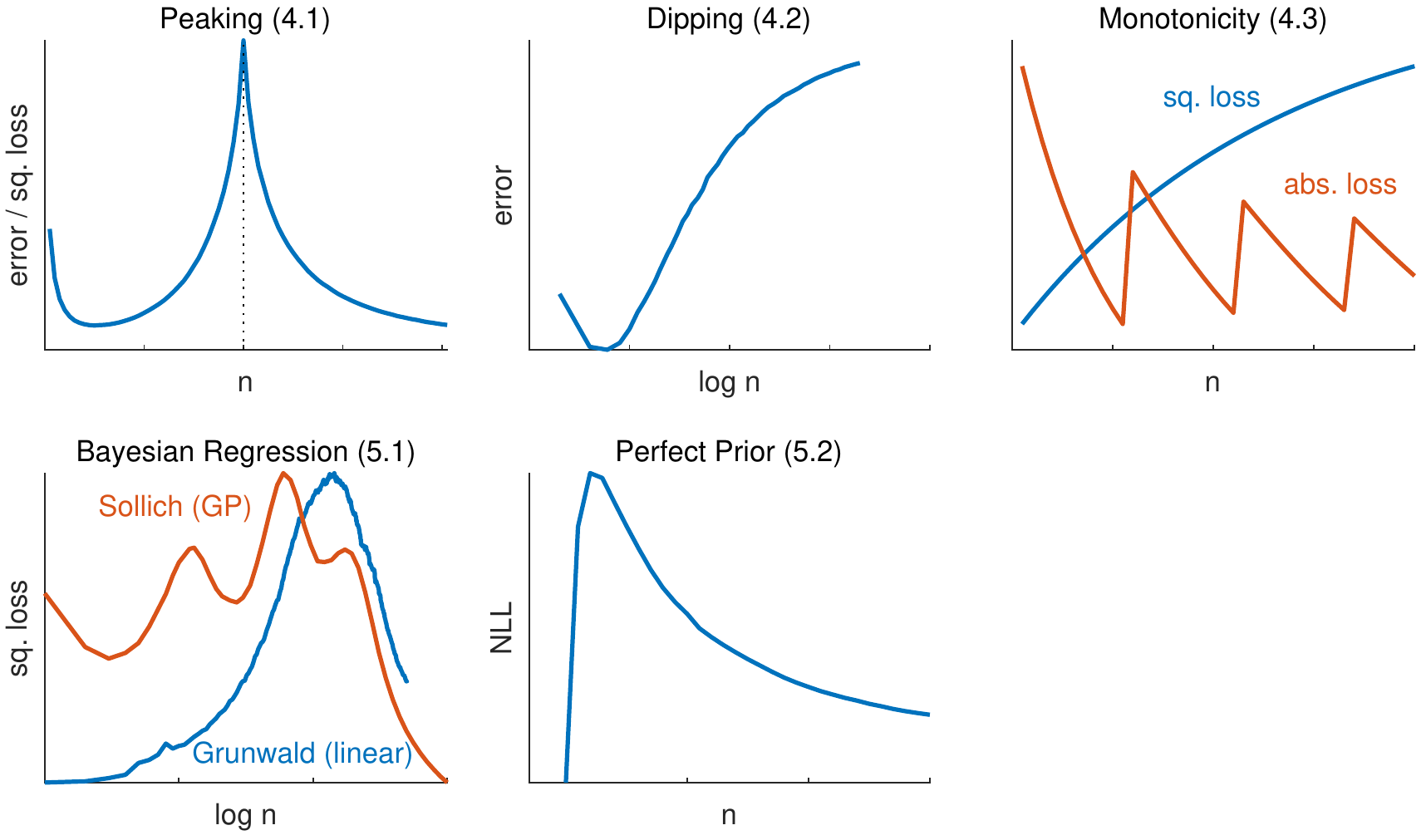}
 	\caption{Qualitative overview of various learning curve shapes placed in different categories with references to their corresponding subsections in the subtitle. All have the sample size $n$ or $\log n$ on the horizontal axis. %
 	Dotted lines indicate the transition from under to overparametrized models. Abbreviations; error: classification error; sq.~loss: squared loss; NLL: negative log likelihood; abs.~loss: absolute loss.}
 	\label{fig:overview}
 \end{figure}

The three subsections cover the three main settings in which bad behavior for standard learning curves can arise. Next to that, we point out what we currently understand of what are the essential differences between these settings.  Subsequently, Section \ref{sect:pabad} turns to PA learning curves. An overview of the qualitative shapes of these learning curves, including the PA ones, is given in Figure \ref{fig:overview}. 

\subsection{Peaking and Double Descent}\label{sub:peaking}

Probably the earliest abnormal behavior identified and studied among learning curves is so-called peaking.  The term indicates that the learning curve takes on a maximum, typically in the form of a cusp. See Figure \ref{fig:overview} (top left). 

Unlike many other bad behaviors, peaking can occur in the realizable setting, i.e., where the learning model is actually well-specified, in the sense that the overall best-performing model is in the model class considered. Its cause seems related to instability of the model, which is maximal around the point where $N$ hits the capacity of the learner.  This phenomenon should not be confused with peaking for feature curves\footnote{A feature curve plots the performance of a learner against the varying number $d$ of measurements it is trained on \cite{hughes1968mean,Jain1982}.} (there are, however, direct connections between these viewpoints  \cite{Duin2000,loog2020brief}).  The feature-curve phenomenon has gained quite some renewed attention in recent years under the name double descent \cite{Belkin2019}.  By now, the term (sample-wise) double descent has become a term for the  peak in the learning curve for deep neural networks as well \cite{nakkiran2019deep,Nakkiran2019more}.

Peaking was first observed for pseudo-Fisher's linear discriminant (PFLD) in 1989 \cite{vallet1989linear} and has been (re)considered at numerous occasions \cite{loog2020brief}. The PFLD is the classifier minimizing the squared loss, using minimum-norm or ridgeless linear regression based on the pseudo-inverse. PFLD often peaks at $d \approx N$, both for the squared loss and  classification error.  A first theoretical model explaining this behavior in the so-called thermodynamical limit is given in \cite{opper1990ability}. In such works, often originating from statistical physics, the usual quantity of interest is $\alpha = \tfrac{d}{N}$ that controls the relative sizes for $d$ and $N$ going to infinity  \cite{opper1991calculation,watkin1993statistical,engel2001statistical}.

Reference \cite{Raudys1998} investigates peaking in the finite sample setting where each class is a Gaussian. They approximately decompose the generalization error in three terms. The first term measures the quality of the estimated means and the second the effect of reducing the dimensionality due to the pseudo-inverse. These terms reduce the error when $N$ increases. The third term measures the quality of the estimated eigenvalues of the covariance matrix. This term increases the error when $N$ increases, because more eigenvalues need to be estimated at the same time if $N$ grows, reducing the quality of their overall estimation. These eigenvalues are often small and as the model depends on their inverse, small estimation errors can have a large effect, leading to a large instability \cite{Skurichina1996} and peak in the learning curve around $N \approx d$.  Using an analysis similar to \cite{Raudys1998}, \cite{Krijthe2016} study the peaking phenomenon in semi-supervised learning and show that unlabeled data can both mitigate or worsen it.

The work in \cite{Duin2000} illustrates experimentally that SVMs may not suffer from peaking and Opper \cite{Opper2001} presents a similar conclusion.  For specific learning problems,  \cite{opper1990ability} and \cite{Watkin1993} already give a theoretical underpinning for the absence of double descent for the perceptron of optimal (or maximal) stability, which is a classifier closely related to the SVM.  \cite{opper2001universal} studies the behavior of the SVM in the thermodynamic limit, which does not show peaking either. \cite{spigler2019jamming} demonstrates, however, that double descent for feature curves can occur using the (squared) hinge loss, where the peak is typically located at a training sample size $N$ that is larger than the dimensionality $d$.

Further insight into when peaking can occur may be obtained from \cite{advani2017high} and  \cite{hastie2019surprises}.  These perform a rigorous analysis of the PFLD and standard least-squares regression using random matrix theory.  Results should, however, be interpreted with care as these are typically derived in an asymptotic setting where both $N$ and $d$ (or some more appropriate measure of complexity) go to infinity, i.e., a setting similar to the earlier mentioned thermodynamic limit. \cite{d2020triple} shows that a peak can occur both where the training set size $N$ equals the input dimensionality $d$ and when $N$ matches the number of parameters of the learner.  This depends on the learner's degree of nonlinearity.  Multiple peaks are also possible for $N<d$  \cite{nakkiran2020optimal}.

\subsection{Dipping and Objective Mismatch}\label{sect:dip}

In peaking, the performance temporarily deteriorates, but recovers with the further increase of $N$. Optimal performance is typically achieved with an infinite sample size.  In dipping, however, the learning curve may initially improve with more samples, but the performance eventually deteriorates and never recovers, even in the limit of infinite data \cite{Loog2012}. Thus the best expected performance is reached at a finite training set size. See Figure \ref{fig:overview} (top middle).

By constructing an explicit problem, \cite{Devroye1996} already showed that the nearest neighbor classifier is not always, what they refer to as, smart, meaning its learning curve can go up locally.  A similar claim is made for kernel rules in Problems 6.14 and 6.15 from \cite{Devroye1996}.

A one-dimensional toy problem for which many well-known linear classifiers dip is easily constructed \cite{Loog2012}.  In a different context, \cite{Ben-David2012} provides an even stronger example where all linear classifiers optimizing a convex surrogate loss converge in the limit to the worst possible classifier for which the error rate comes arbitrarily close to 1. Another example, Lemma 15.1 in \cite{Devroye1996}, gives an very simple case of dipping for the likelihood where merely the estimation of the a prior class probability is considered.

What is essential for dipping to occur is that the the hypothesis class is misspecified and that the learner optimizes something else than the evaluation metric of the learning curve.  Such objective misspecification is standard since many evaluation measures such as error rate, AUC, F-measure, and so on, are notoriously hard to optimize (see, e.g., \cite{shalev2014understanding}).   In all of the above examples, the evaluation measure was the 0-1 loss, but the classifiers were optimized based on some standard surrogate measure like the hinge loss, the squared loss, or the likelihood.

Other works also show dipping of some sort. For example, \cite{Frey1999} fits C4.5 to a synthetic dataset that has binary features for which the parity of all features determines the label. When fitting C4.5 the test error increases with the amount of training samples. They attribute this to the fact that the C4.5 is using a greedy approach to minimize the error and, as such, is closely related to objective misspecification.   \cite{Brumen2012} shows a badly behaving curve for C4.5 that goes up. In addition, another 34 other curves were reported to not fit well using their parametric models, which may point to similar problems of curve increase. In \cite{vanschoren2008learning}, we find another potential example of dipping as, in Figure 6, the accuracy goes down with increasing sample sizes. 

Anomaly or outlier detection using $k$-nearest neighbors ($k$NN) can shows dipping behavior as well \cite{Ting2017}. Also here is a mismatch between the objective that is evaluated with, i.e., the AUC and $k$NN that does not optimize the AUC. \cite{Hess2010} already shows $k$NN learning curves that deteriorate in terms of AUC in the standard supervised setting. 

In active learning for classification, where the test error rate is often plotted against the size of the (actively sampled) training set, learning curves are regularly reported to dip  \cite{schohn2000less,konyushkova2015introducing}.  That is, active learners provide optimal performance for a number of labeled samples that is smaller than the complete training set. Possibly, the active learner merely beats the inactive learner because it uses an objective that better matches the evaluation measure employed \cite{loog2016empirical}.  Finally, so-called negative transfer \cite{wang2019characterizing}, as it occurs in transfer learning and domain adaptation, can be interpreted as dipping as well. In this case, more source data deteriorates performance on the target and the objective mismatch stems from the combined training from source and target data instead of the latter only.

\subsection{Risk Monotonicity and Empirical Risk}\label{sect:mono}

Several novel examples of nonmonotonic behavior for density estimation, classification, and regression by means of standard empirical risk minimization (ERM) are shown in \cite{loog2019minimizers}. Similar to dipping, at some point, the squared loss increases with increasing $N$, but, in contrast with dipping, does eventually recover. See Figure \ref{fig:overview} (top right).

These examples can neither be explained in terms of dipping nor in terms of peaking.  Dipping is ruled out as, in ERM, the learner actually optimizes the loss that is used for evaluation.  In addition, it is shown that learning problems can be constructed such that they act nonmonotonically at any sample size $N$.  As nonmonotonicity can occur at small and large sample sizes, there is no link with the capacity of the learner and we can  rule out an explanation in terms of peaking.

Reference \cite{loog2019minimizers} proofs nonmonotonicity for squared, absolute, and hinge loss.  It demonstrates that likelihood estimators suffer the same deficiency. Two learners are reported that are provably monotonic: mean estimation based on the (negative) log-likelihood and the memorize algorithm from \cite{shalev2014understanding}. The latter algorithm does not really learn but outputs the majority voted classification label of each object if it has been seen before. Memorize is not PAC learnable \cite{shalev2014understanding,engel2001statistical}, illustrating that monotonicity and PAC are, in that sense, different concepts.

The work in \cite{Viering2019} and \cite{loog2019minimizers} shows experimentally that regularization can actually worsen the nonmonotonic behavior.  This possibility had already been pointed out in \cite{Grunwald2011} (see Subsection \ref{sec:perfect-prior}).  Another experiment in \cite{loog2019minimizers} shows a surprisingly jagged learning curve for the absolute loss,  Recent work \cite{chen2022} explains both behaviors and shows that the latter curves goes up and down perpetually.

\section{PA Curves and Bad Behavior}\label{sect:pabad}

Where standard learning curves deal with a single learning problem $P$, PA learning curves report an averaged learning curve, where the averaging is done according to a distribution $\mathcal{P}$ over learning problems.  Using Bayesian inference, the PA learning curve is monotonic if the assumed prior over learning problems is correct, i.e., if the prior equals $\mathcal{P}$.  This is a consequence of the total evidence theorem \cite{savage1954foundations,Good1967}, which states, informally, that one obtains the maximum expected utility by taking into account all observations.  As soon as there is any form of misspecification, also badly behaving PA curves can appear. See Figure \ref{fig:overview} (bottom) for an overview of badly behaving PA curves. It is primarily the Bayesian regression setting that has been investigated.

\subsection{Misspecified Bayesian Regression}\label{sub_misspecified_PA}

Gaussian process regression, a particular instance of Bayesian regression, has been studied where the so-called teacher model provides the training data, while the student model learns, assuming a covariance or noise model different from the teacher.  \cite{Sollich2002b} analyzes the PA learning curve based on the eigenvalue decomposition of the covariances underlying the student and teacher model. The paper assumes both student and teacher use kernels with the same eigenfunctions but possibly differing eigenvalues. Subsequently, it considers various synthetic distributions for which the eigenfunctions and eigenvalues can be computed analytically and finds that for a uniform distribution on the vertices of a hypercube, multiple overfitting maxima and plateaus may be present in the learning curve. See Figure \ref{fig:overview} (bottom left, red) for an example based on an actual experiment.
For a uniform distribution in one dimension, the claim is that there may be arbitrarily many overfitting maxima.

The work in \cite{Grunwald2017} shows that a (hierarchical) Bayesian linear regression model can give a broad peak in the learning curve of the squared risk. See Figure \ref{fig:overview} (bottom left, blue) for one of their curves. One way this can happen is when the homogeneous noise assumption is violated, while the estimator is otherwise consistent. Specifically, let data be generated as follows. For each sample, a fair coin is flipped. Heads means the sample is generated according to the ground truth probabilistic model contained in the hypothesis class. Misspecification happens when the coin comes up tails and a sample is generated in a fixed location without noise. The peak in the learning curve cannot be explained by dipping, peaking or known sensitivity of the squared loss to outliers. The peak in the learning curve is fairly broad and occurs in various experiments. As also no approximations are to blame, \cite{Grunwald2017} concludes that Bayes' rule is really at fault as it cannot handle the misspecification.

\subsection{The Perfect Prior} \label{sec:perfect-prior}

A monotonic PA curve does not rule out that the learning curve for individual problems can go up, even if the problem is well-specified.  \cite{Grunwald2011} offer an insightful example:
consider a fair coin and let us estimate its probability $p$ of heads using Bayes' rule. We measure performance using the negative log-likelihood on an unseen coin flip and adopt a uniform Beta(1,1) prior on $p$. %
This prior, i.e., without any training samples, already achieves the optimal loss since it assigns the same probability to heads and tails. After a single flip, $N=1$, the posterior is updated and leads to a probabilities of $\tfrac{1}{3}$ or $\tfrac{2}{3}$ and the loss must increase. Eventually, with $N$ going to infinity, the optimal loss is recovered, forming a bump in the learning curve. See Figure \ref{fig:overview} (bottom middle) for the resulting curve. Note that this construction is rather versatile and can create nonmonotonic behavior for practically any Bayesian estimation task.  In a similar way, any type of regularization can lead to comparable learning curve shapes, as was already mentioned in Subsection \ref{sect:mono}  (see also  \cite{Viering2019,loog2019minimizers}).

A related example can be found in \cite{Al-Saleh2003}. It shows that the posterior variance can also increase for a single problem, unless the likelihood belongs to the exponential family and a conjugate prior is used. Gaussian processes  fall in this last class and, as such, their PA curve is monotone if there is no model misspecification.

\section{Fixing Monotonicity}\label{sect:fix}

Some works set out to restore monotonicity in rather specific settings.  Peaking of the PFLD, as defined and discussed in Subsection \ref{sub:peaking}, can be avoided through regularization, though the tuning has to be done carefully \cite{Raudys1998,Skurichina1996,Tresp2005,Skurichina1996}. Assuming the data is isotropic, \cite{nakkiran2020optimal} shows that peaking disappears for the optimal setting of the regularization parameter. Other, more heuristic solutions change the training procedure altogether, e.g., \cite{Duin1995} uses an iterative procedure that decides which objects PFLD should be trained on. \cite{Skurichina1999} adds copies of objects with noise or increases the dimensionality by adding noise features. 

Reference \cite{Sollich2004} extends the study of Gaussian process regression from Subsection \ref{sub_misspecified_PA} and suggests to optimize hyperparameters such as length scale and noise level during learning based on evidence maximization.  Among others, the paper finds that the earlier considered hypercube does not lead to arbitrary many overfitting maxima anymore.  In fact, the learning curve becomes monotonic.  Following the Bayesian regression analysis of \cite{Grunwald2017}, as covered in Subsection \ref{sub_misspecified_PA}, the same work introduces a modified Bayes rule, where the likelihood is raised to some power.  This parameter, however, cannot be learned in a Bayesian way, leading to their safe-Bayes approach. This technique alleviates the broad peak in the learning curve and is empirically shown to make the curves more well-behaved. 

A first attempt at a more generally applicable approach to restore monotonicity, though focusing on 0-1 loss, is made by \cite{Viering2020}. They propose a wrapper that, with high probability, makes any classifier monotonic in terms of the the error rate. The main idea is to consider $N$ as a variable over which model selection is performed. When $N$ is increased, a model trained with more data is compared to the previously best model on validation data. Only if the new model is judged to be significantly better, the older model is discarded. If the original learning algorithm is consistent and if the size of the validation data grows, the resulting algorithm is consistent as well. It is empirically observed that the monotonic version may learn slower.

This idea is extended in \cite{mhammedi2021risk}, which proposes two algorithms that do not need to set aside validation data while guaranteeing monotonicity. To this end they assume that the Rademacher complexity of the hypothesis class composited with the loss is finite. This allows it to determine when to switch to a model trained with more data. In contrast to \cite{Viering2020}, it argues that the second algorithm does not learn slower, as its generalization bound coincides with a known lower bound of regular supervised learning.

Finally, \cite{bousquet2022monotone} offers an analysis of the intricacies that can play a role in the potentially bad behavior of learning curves under the 0-1 loss. Based on these insights they formulate a general, sophisticated transformation applicable to a base learner that leads to monotonic behavior.  The uniform error bounds they provide for their approach readily implies that every PAC learnable class admits a monotonic learner.  Notably, their work took inspiration from \cite{pestov2021universally} in which it is shown that universally consistent (binary) classifiers can be monotonic (notably, it gives an explicit method to construct a monotonic histogram classifiers).  This last, actually rather recent result disproves a conjecture from \cite{Devroye1996} that posits that such universally consistent classifiers do not exist.

\section{Discussion and Conclusions}\label{sect:disc}

In the past few decades, it are especially the more theoretical papers that have studied badly behaving learning curves. In addition, it seems that the monotonicity problem is currently gaining some traction from that side of the research spectrum as well.  In our personal experience, many practitioners dismiss concerns over nonmonotonic behavior with the remark that such is only found in rather artificial settings.  Often, the claim is made that such behavior does not occur in real-world applications or on real-world data.

But how do we know?  Even if it is the case that this behavior does not happen in practice, this is still left to be demonstrated.  Besides, the relatively artificial settings that some proofs may rely on, at the very least, show that there is something we do not understand.  Therefore, these examples should be taken as a starting point to improve our understanding of the learning process.  We should try to understand, both in practice and theory, how bad this behavior can get, how it can be mitigated, and what we gain or loose by this.  Such understanding may certainly be of direct relevance if we want to exploit the usage of learning curves, as covered in Section \ref{sect:use}, to its full extent.  In all, empirical and theoretical research challenges abound (see also \cite{bousquet2022monotone,loog2019minimizers,mohr2022lcdb} for instance).

\subsection{Nonmonotonicity in Other AI Disciplines}

This survey's focus is on standard supervised learning, but what can we say about more complicated learning settings?   What to expect, for instance, when dealing with adversaries, when we act in an environment that is only partially observable, or if we are dealing with multiple agents?  Arguably, we can expect even more exotic learning behavior in general and, therefore, also new settings in which learning can uniquely fail.

\subsection{Meta-Learning}

We think better understanding is needed with respect to the occurrence of peaking, dipping, and otherwise nonmonotonic behavior.   As indicated already, the simple fact is that, at this point, we do not know what role these phenomena play in real-world problems.   Now that many benchmark datasets are readily available, this issue can be studied more rigorously.  Properly summarizing and openly sharing learning curve data can further support this research. Automated techniques may then be developed to find curious learning curve phenomena and possibly predict them.

Given the success of meta-learning for curve extrapolation and model selection \cite{Leite2004,strang2018don} this seems a promising possibility. Such meta-learning studies on large amounts of datasets could, in addition, shed more light on what determines the parameters of learning curve models, a topic that has been investigated relatively little up to now. Predicting these parameters robustly from very few points along the learning curve will prove valuable for virtually all applications.

An initial study into the behavior of learning curves on a newly developed database can be found in \cite{mohr2022lcdb}.  The database makes learning curves for 246 datasets and 20 classifiers publicly available and presents some basic and preliminary findings.  Among them, the most interesting from a practical point of view may be that peaking does not seem to occur very often.  In all, this database offers ample opportunity to investigate learning curves extensively in an empirical way.

\subsection{Two Theoretical Questions}

There are two rather specific theoretical questions that we like readers to consider.

The first one asks whether maximum likelihood estimators for well-specified models behave monotonically.  Likelihood estimation, being a century-old, classical technique \cite{fisher1912absolute,stigler2007epic}, has been heavily studied, both theoretically and empirically.  In much of the theory developed, the assumption that one is dealing with a correctly specified model is common, but we are not aware of any results that demonstrate that better models are obtained with more data.  The question is interesting for the likelihood exactly because this estimator has been extensively studied already and still plays a central role in statistics and abutting fields as well.

The second question is broader: for standard regression problems, among the consistent learners are there monotonic ones?  \cite{bousquet2022monotone} demonstrates that for the classification setting monotonic learners can be designed and they ask the same question for more general loss function.   They indicate, however, that one should probably restrict oneself to bounded losses.  A derivative question could therefore be: for unbounded losses, can we show that for every (consistent) learners a problem exists on which this learner behaves badly.

\subsection{Concluding}

It is striking that there is still so much that we actually do not understand about learning curve behavior and, as such, learning itself. Even some of the most simple settings elude us.  Most theoretical results are restricted to relatively basic learners, while the empirical research that has been carried out is rather limited in scope.  We identified some specific challenges in the foregoing, but we are convinced that many more interesting problems can be discovered.  We are convinced that a deeper understanding of badly behaving learning curves will also turn out to be practically beneficial.  To us, however, a sufficiently valid reason to investigate it should be to quench one's scientific curiosity.

\bibliographystyle{splncs04}
\bibliography{bib}

\end{document}